\documentclass[conference]{IEEEtran}
\IEEEoverridecommandlockouts
\usepackage{cite}
\usepackage{amsmath,amssymb,amsfonts}
\usepackage{graphicx}
\usepackage{textcomp}
\usepackage{xcolor}
\usepackage{url}
\usepackage{subcaption}
\usepackage{algorithm}
\usepackage{algpseudocode}
\usepackage{hyperref}
\usepackage{listings}
\usepackage{tcolorbox}

\usepackage{authblk}

\def\BibTeX{{\rm B\kern-.05em{\sc i\kern-.025em b}\kern-.08em
    T\kern-.1667em\lower.7ex\hbox{E}\kern-.125emX}}
\begin{document}

% \title{Multi-Agent Interaction and Decision Making in LLMs through Ô-Ăn-Quan Game\\
% \title{Multi-Agent Interaction and Decision Making through Traditional Vietnamese Board Game\\
% \title{Ô Ăn Quan Game: Multi-Agent Interaction and Decision Making in LLMs\\
%\title{Can LLMs Play "Ô Ăn Quan" ? \\
%Multi-Step Planning and Decision Making\\
%}

% \title{Can LLMs Play Ô Ăn Quan Game? \\ A Study of Multi-Step Planning and Decision Making}
% \title{\parbox{\linewidth}{\centering Can LLMs Play Ô Ăn Quan Game? A Study of Multi-Step Planning and Decision Making}}

\title{\fontsize{22pt}{22pt}\selectfont Can LLMs Play Ô Ăn Quan? \\ A Study of Multi-Step Planning and Decision Making}

\author{
Sang Quang Nguyen$^{1,2}$, 
Kiet Van Nguyen$^{1,2}$, 
Vinh-Tiep Nguyen$^{1,2}$, \\
Thanh Duc Ngo$^{1,2}$,
Ngan Luu-Thuy Nguyen$^{1,2}$, 
Duy-Dinh Le$^{1,2}$ \\
\small $^1$University of Information Technology, Ho Chi Minh City, Vietnam \\
\small $^2$Vietnam National University, Ho Chi Minh City, Vietnam \\
\small \text{sangnq.19@grad.uit.edu.vn} \\
\small \text{\{kietnv, tiepnv, thanhnd, ngannlt, duyld\}@uit.edu.vn}
}

\maketitle

\begin{abstract}
In this paper, we explore the ability of large language models (LLMs) to plan and make decisions through the lens of the traditional Vietnamese board game, Ô Ăn Quan. This game, which involves a series of strategic token movements and captures, offers a unique environment for evaluating the decision-making and strategic capabilities of LLMs. Specifically, we develop various agent personas, ranging from aggressive to defensive, and employ the Ô Ăn Quan game as a testbed for assessing LLM performance across different strategies. Through experimentation with models like Llama-3.2-3B-Instruct, Llama-3.1-8B-Instruct, and Llama-3.3-70B-Instruct, we aim to understand how these models execute strategic decision-making, plan moves, and manage dynamic game states. The results will offer insights into the strengths and weaknesses of LLMs in terms of reasoning and strategy, contributing to a deeper understanding of their general capabilities.

\end{abstract}


\section{Introduction}

Large Language Models (LLMs) such as GPT~\cite{brown2020language}, Llama~\cite{touvron2023llama, touvron2023llama2, grattafiori2024llama}, and Gemini~\cite{team2023gemini} have demonstrated remarkable performance across various natural language understanding and generation tasks. While their capabilities in handling tasks like translation~\cite{zhang2023prompting, kocmi2023large}, summarization~\cite{zhang2024comprehensive}, and question answering~\cite{lewis2020retrieval} have been extensively studied. Most existing game testbed focus on imperfect-information games (e.g., Werewolf~\cite{huang2024far, sato2024implementation} , Avalon~\cite{lan2023llm}, Spyfall~\cite{wei2025exploring}) or games with heavy reliance on deception and bluffing. In contrast, Ô Ăn Quan offers a unique structure of fully observable, deterministic turn-based gameplay with circular movement and resource looping, which presents a new challenge for LLMs in reasoning under structured progression.

\begin{figure*}[h]
\centering
\includegraphics[width=0.99\textwidth]{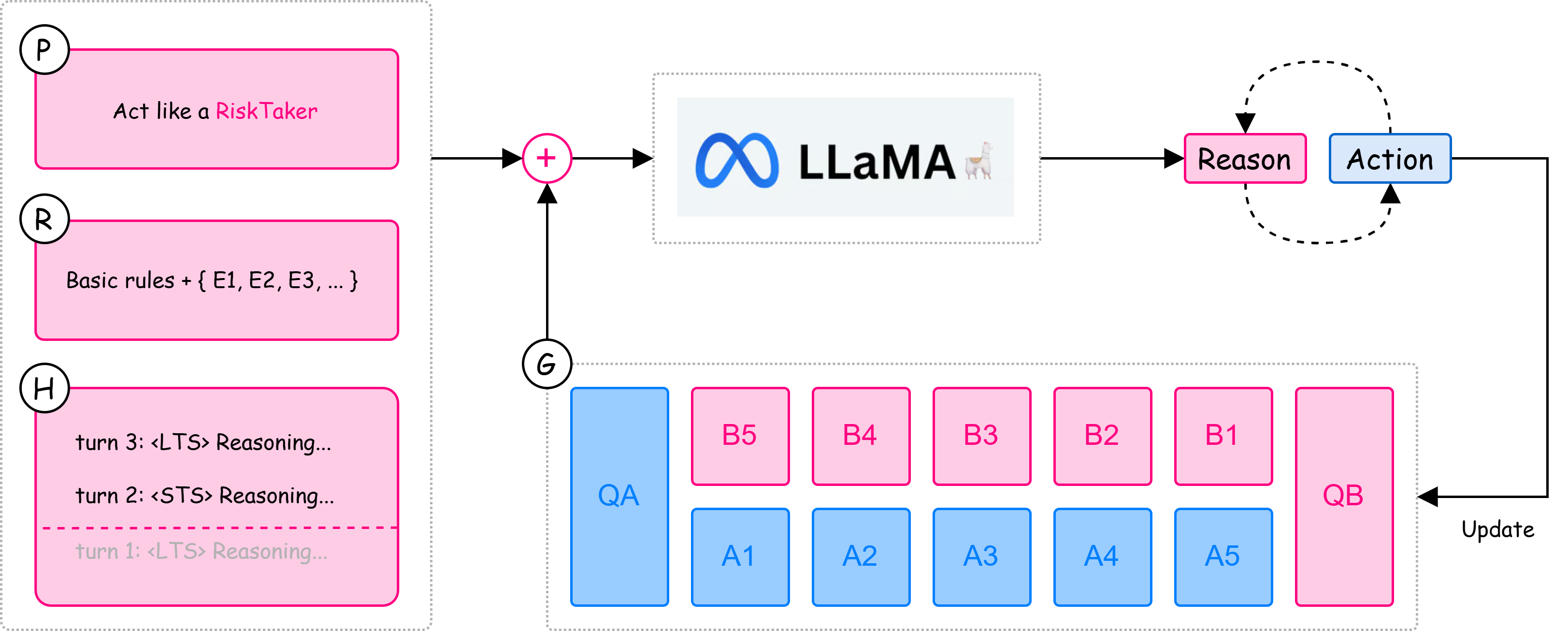}
\caption{Overview of the LLM-based agent framework for \textit{Ô Ăn Quan}. The system takes as input the current game state $\mathcal{G}$, reasoning history $\mathcal{H}$, and rule instructions $\mathcal{R}$, then generates a natural language rationale and action via a LLaMA-based model. The predicted action updates the board, and the process continues turn-by-turn in a closed-loop interaction.}
\label{fig:overview-framework}
\end{figure*}

Ô Ăn Quan~\footnote{https://en.wikipedia.org/wiki/Ô\_ăn\_quan} is played between two players who take turns distributing tokens across the board and capturing tokens from their opponent. The game involves strategic planning, including managing resources and controlling key areas (referred to as the "Quan"). In the context of LLMs, we can model different agent types (personas) to simulate different strategies and evaluate the LLM's ability to plan, reason, and adapt in a dynamic environment. By leveraging the rules and structure of Ô Ăn Quan, we can analyze how LLMs make decisions based on the game state, evaluate their ability to handle both short-term and long-term planning, and determine their overall strategic thinking.

This paper aims to bridge this gap by designing a series of experiments using different LLMs, ranging from smaller models such as Llama-3.2-3B-Instruct to larger counterparts such as Llama-3.1-8B-Instruct and Llama-3.3-70B-Instruct. The objective is to evaluate their performance in strategic multi-agent settings, specifically focusing on their ability to engage in the Ô Ăn Quan game—a traditional Vietnamese board game that combines elements of turn-based strategy, resource management, and foresight. Additionally, the study will examine how the adoption of diverse personas influences model decision-making and assess whether these models can adapt their strategies in response to dynamic game environments.

\section{Related Work}

Decision-making in interactive environments has been a significant area of research in both artificial intelligence (AI) and game theory. Previous works have explored various methods for modeling and evaluating strategic decision making, from traditional game theory~\cite{von2007theory} to modern reinforcement learning (RL) and adversarial AI strategies~\cite{silver2016mastering}. In particular, recent studies have used board games such as Chess~\cite{silver2018general} and Go~\cite{silver2017mastering} as benchmarks to evaluate the performance of AI systems in complex and dynamic environments that require planning and long-term strategy.

In the realm of natural language processing (NLP), LLMs have been extensively tested for tasks involving decision-making, reasoning, and planning. For example, studies such as those of~\cite{brown2020language} and \cite{rabinowitz2018machine} have demonstrated LLM capabilities in structured planning environments and strategic dialogue systems. These works underscore the importance of evaluating language models not only for their ability to answer factual questions but also for their capability to handle open-ended, multi-step tasks that require strategic reasoning.

However, to date, there has been limited work that applies LLMs in interactive games that require both strategic decision-making and dynamic adaptability. Our approach is inspired by the growing interest in using games as testing ground~\cite{huang2024far, sato2024implementation, lan2023llm, wei2025exploring}. Specifically, we focus on Ô Ăn Quan, a traditional Vietnamese game with a rich set of rules that offers a novel domain to test the limits of LLM planning abilities.

\section{Game Environment Implementation}

\textit{Ô Ăn Quan} is a traditional Vietnamese board game played between two players on a 2x6 grid as shown in Figure~\ref{fig:overview-framework}. Each player controls five regular positions and a central \textit{Mandarin} position (\textit{Quan}) on their side of the board. The objective is to maximize one's score by capturing as many tokens as possible, including the opponent’s Mandarin.

% \begin{figure}[h]
% \centering
% \includegraphics[width=0.49\textwidth]{figures/game-board2.png}
% \caption{Ô Ăn Quan game board.}
% \label{fig:game-board}
% \end{figure}

\begin{figure*}[ht]
    \centering
    \includegraphics[width=0.98\textwidth]{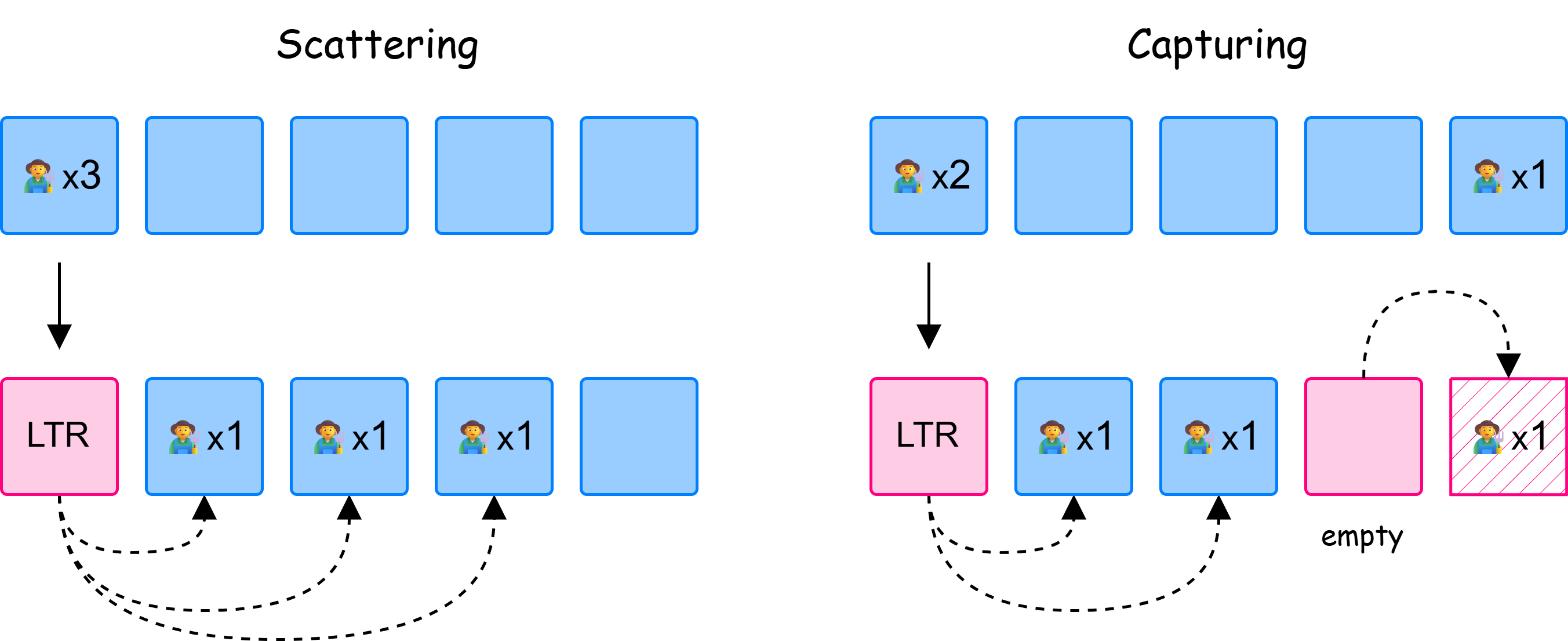}
    \caption{Illustration of token movement in Ô Ăn Quan: from scattering to capturing.}
    \label{fig:scattering_capturing}
\end{figure*}

Gameplay consists of two alternating phases as shown in Figure~\ref{fig:scattering_capturing}: the \textbf{Scattering Phase}, where a player selects one of their positions and redistributes all peasant tokens it contains clockwise or counterclockwise around the board; and the \textbf{Capturing Phase}, where tokens may be captured based on specific spatial configurations. The game concludes when both Mandarin tokens are captured or no legal moves remain.

Beyond its simple rules, Ô Ăn Quan presents a rich set of strategic challenges. Players must engage in \text{strategic planning}, carefully selecting where to begin scattering in order to chain moves, set traps, and avoid future vulnerabilities. Effective \text{resource management} is also critical, as aggressive captures must be weighed against the need to maintain sufficient tokens to sustain future turns. Finally, the game demands high \text{adaptability}, as the board state evolves dynamically with each move, requiring continuous reassessment of threats and opportunities.

% \subsection{Key Challenges}

% The game presents several strategic challenges:

% \begin{itemize}
%     \item \textbf{Strategic Planning:} Players must carefully choose where to begin scattering, aiming to chain moves and set traps for capturing, while avoiding vulnerabilities.

%     \item \textbf{Resource Management:} Players must balance aggressive captures with maintaining enough tokens to continue playing, since empty sides can lead to forced redistribution.

%     \item \textbf{Adaptability:} The board state changes rapidly, requiring players to revise strategies based on evolving threats and opportunities.

% \end{itemize}

\subsection{Extra Rules: Strategic Challenges for LLMs}

% To further evaluate the planning capabilities of large language models (LLMs), we introduce several rule extensions that increase strategic complexity.

\textbf{E1 – Immature Mandarin Rule.} A Mandarin with fewer than 5 tokens (also referred to as \textit{Quan Non}) cannot be captured.

\textbf{E2 – Forced Redistribution.} If a player has no peasants remaining in their positions, they must return 5 captured tokens to the board to continue the game.

\textbf{E3 – Early Game Restriction.} No Mandarin captures are allowed during the first two rounds, encouraging early-game setup rather than aggression.

\textbf{E4 – Two-Empty Rule.} If the two positions immediately beyond the last scattered token are empty or invalid, the player’s turn ends immediately.

\textbf{E5 – Forced Capture Chain.} Whenever a valid capture is available, it must be executed, and the player must continue capturing as long as additional valid captures exist.

These additional rules act as stress tests for LLM agents, requiring them to exhibit deeper planning, enforce rule compliance, and move beyond simplistic or greedy strategies.

\subsection{Core Actions: Scattering and Capturing}

\begin{algorithm}[ht]
\caption{Basic Scattering}
\begin{algorithmic}[1]
\Require Board state $B[0 \ldots N-1]$, selected index $i$, direction $D \in \{\text{LTR}, \text{RTL}\}$
\State $T \gets B[i]$
\For{$k = 1$ to $T$}
    \If{$D = \text{LTR}$}
        \State $j \gets (i + k) \bmod N$
    \Else
        \State $j \gets (i - k + N) \bmod N$
    \EndIf
    \State $B[j] \gets B[j] + 1$
\EndFor
\State $B[i] \gets 0$
\end{algorithmic}
\label{alg:scattering}
\end{algorithm}

\textbf{Scattering} is the process by which a player redistributes peasant tokens across the board from a chosen starting position. As shown in Algorithm~\ref{alg:scattering}, the player selects a direction and distributes all tokens from the chosen position one by one into subsequent positions, wrapping around the board as needed.

\begin{algorithm}[ht]
\caption{Basic Capturing}
\begin{algorithmic}[1]
\Require Final scatter index $i$, direction $D \in \{\text{LTR}, \text{RTL}\}$, board state $B[0 \ldots N-1]$
\State $\text{Captured} \gets 0$
\While{True}
    \If{$D = \text{LTR}$}
        \State $j_1 \gets (i + 1) \bmod N$
        \State $j_2 \gets (i + 2) \bmod N$
    \Else
        \State $j_1 \gets (i - 1 + N) \bmod N$
        \State $j_2 \gets (i - 2 + N) \bmod N$
    \EndIf
    \If{$B[j_1] = 0$ \textbf{and} $B[j_2] > 0$}
        \State $\text{Captured} \gets \text{Captured} + B[j_2]$
        \State $B[j_2] \gets 0$
        \State $i \gets j_2$
    \Else
        \State \textbf{break}
    \EndIf
\EndWhile
\end{algorithmic}
\label{alg:basic_capturing}
\end{algorithm}

\textbf{Capturing} occurs immediately after scattering, as described in Algorithm~\ref{alg:basic_capturing}. If the next position in the given direction is empty, and the position beyond it contains tokens, those tokens can be captured. This process may repeat recursively as long as the condition remains valid.

\section{Agent Implementation}

To evaluate the strategic reasoning capabilities of large language models (LLMs), we implement agents based on the \text{Llama-3.2-3B}, \text{Llama-3.1-8B}, and \text{Llama-3.3-70B} families. Each agent is assigned a behavioral \textit{persona}, described in natural language, such as aggressive, defensive, greedy, or balanced. These personas influence the agent's reasoning pattern and decision-making process throughout the game.

\begin{equation}
\text{LLM}_\theta(\mathcal{G},\ \mathcal{H},\ \mathcal{R},\ \mathcal{P}) \rightarrow (reason,\ action)
\label{eq:agent_input_output}
\end{equation}

In equation \ref{eq:agent_input_output}, $\mathcal{G}$ represents the current board state, while $\mathcal{H}$ provides a textual summary of the previous turn, including the opponent’s move and rationale. The term $\mathcal{R}$ denotes the rule specification, encompassing both basic and extended game rules. $\mathcal{P}$ encodes the agent’s persona as a natural language instruction. The model outputs a natural language explanation \textit{reason}, and an \textit{action} that includes both the selected starting position $B[i]$ and the movement direction $D$. The agent's decision process is defined as \ref{eq:agent_input_output}, this formulation follows recent prompting strategies \cite{wei2022chain, yao2023react} and persona-based agent modeling \cite{park2023generative}, enabling a consistent and interpretable framework to probe how different LLMs make decisions in a two-agent, turn-based setting.

\section{Experiments and Discussion}

We design a series of experiments to evaluate the planning, decision-making, and rule-following behavior of language models in the Ô Ăn Quan environment. Each experiment focuses on a specific aspect of performance, ranging from strategy adherence to rule compliance.

\subsection{Effectiveness by Agent Strategy}

To evaluate how different reasoning styles affect gameplay performance, we conducted a series of controlled experiments in which a fixed baseline \textit{RandomAgent}, an agent that selects legal moves and directions randomly—was used as the opponent. Each persona-based agent (e.g., \textit{BalancedAgent}, \textit{RiskTakerAgent}, etc.) played 50 games against the RandomAgent under identical initial conditions.

% \begin{figure}[h]
% \centering
% \scalebox{0.58}{
% \includegraphics{figures/wr_agent_persona.png}
% }
% \caption{Win/Draw/Loss distribution across 50 games for each agent persona. The dotted line represents the balance point (25 games).}
% \label{fig:wr_agent_persona}
% \end{figure}

% Figure~\ref{fig:wr_agent_persona} further supports these observations: \textit{QuickPlayAgent} achieves the highest win rate with 22/50 wins (44\%), while \textit{BalancedAgent} follows closely with 21/50 wins (42\%). Both also show above-average draw rates (20\% and 18\% respectively). Meanwhile, \textit{RiskTakerAgent}, despite its bold playstyle, only manages 18/50 wins and suffers the highest number of losses.

\begin{table}[h]
\centering
\caption{Average Scores Across Game Phases by Agent Strategy. 
EGE: Early Game End (Rounds 1–10), MGE: Mid Game End (Rounds 11–20), LGE: Late Game End (Rounds 21–25). The \textbf{best} overall results are in bold, and the \underline{second best} are underlined.}
\renewcommand{\arraystretch}{1.3}
\begin{tabular}{|l|c|c|c|c|}
\hline
\textbf{Agent} & \textbf{EGE} & \textbf{MGE} & \textbf{LGE} & \textbf{Avg Point} \\
\hline
BalancedAgent         & 28.5 & \textbf{28.0} & 19.6 & 22.6 \\
DefensiveAgent        & 29.1 & \underline{27.3} & \underline{20.5} & \textbf{23.5} \\
QuickPlayAgent        & \underline{29.7} & 26.0 & 19.7 & 22.4 \\
RiskTakerAgent        & 29.5 & 26.4 & 19.4 & \underline{23.0} \\
StrategicControlAgent & \textbf{30.0} & 25.4 & \textbf{21.3} & 22.5 \\
\hline
\end{tabular}
\label{tab:agent_phase_scores}
\end{table}

Table~\ref{tab:agent_phase_scores} complements these findings by providing a phase-wise breakdown of average scores. 
\textit{StrategicControlAgent} dominates the early game phase (EGE = 30.0), reflecting its emphasis on establishing early control. 
However, it struggles to maintain this advantage into the mid and late game. 
In contrast, \textit{DefensiveAgent} and \textit{BalancedAgent} maintain relatively stable scores throughout, with \textit{DefensiveAgent} achieving the highest average point overall (23.5). 
This reinforces the observation that while early aggression may yield short-term benefits, strategies with better long-term balance tend to perform more consistently across full matches.

\subsection{Comparison Across LLM Architectures}

\begin{table}[h]
\centering
\caption{Win and Draw Rates of LLM Agents in Matches against Llama-3.3-70B-Instruct ($\dagger$ denotes the first player). The \textbf{best} overall results are in bold, and the \underline{second best} are underlined.}
\renewcommand{\arraystretch}{1.3}
\begin{tabular}{|l|c|}
\hline
\textbf{Model (as opponent)} & \textbf{Win (\%) / Draw (\%)} \\
\hline
Llama-3.2-3B-Instruct & 24 / 24 \\
Llama-3.1-8B-Instruct & \textbf{38} / 22 \\
Llama-3.3-70B-Instruct\textsuperscript{$\dagger$} & \underline{34} / 24 \\
\hline
\end{tabular}
\label{tab:agent_win_draw_percent}
\end{table}

Table~\ref{tab:agent_win_draw_percent} presents the win and draw rates of various LLM agents when matched against \text{Llama-3.3-70B-Instruct}, the strongest model in our evaluation. We observe that \text{Llama-3.1-8B-Instruct} achieves the highest win rate (38\%) despite having fewer parameters than the 70B model. This suggests that smaller models may sometimes adopt simpler, more aggressive strategies that exploit early-game advantages or opponent mistakes. The 3B variant, \text{Llama-3.2-3B-Instruct}, shows a balanced performance with a win and draw rate of 24\% each, indicating a tendency toward more neutral, less decisive outcomes—possibly due to limited capacity for long-term planning. Interestingly, when \text{Llama-3.3-70B-Instruct} plays first (as denoted by $\dagger$), it secures a 34\% win rate and a 24\% draw rate. While its larger capacity may contribute to deeper planning and more cautious play, it does not always translate into dominance over smaller models, especially when those models rely on opportunistic tactics. These findings suggest that model scale alone is not a sufficient predictor of success in strategic two-agent settings. Instead, interaction dynamics, first-move advantage, and decision-making patterns play crucial roles in determining performance.

\subsection{Planning Depth Analysis}

\begin{figure*}[ht]
    \centering
    \begin{subfigure}[b]{0.3\textwidth}
        \centering
        \includegraphics[width=\textwidth]{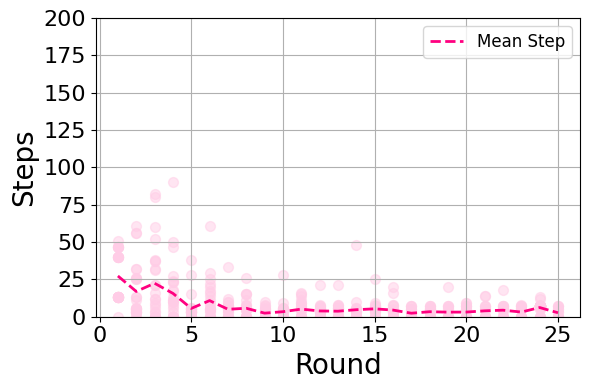}
        \caption{Llama-3.2-3B-Instruct}
        \label{fig:step_3b}
    \end{subfigure}
    \hfill
    \begin{subfigure}[b]{0.3\textwidth}
        \centering
        \includegraphics[width=\textwidth]{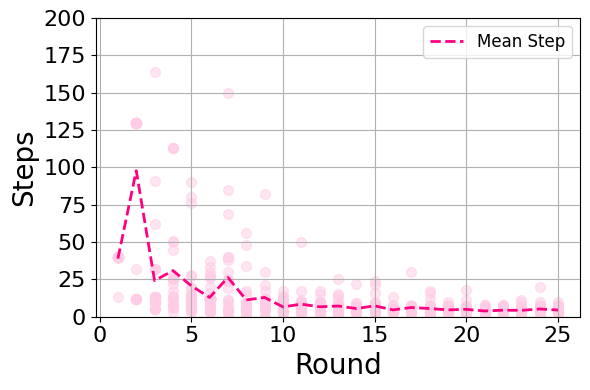}
        \caption{Llama-3.1-8B-Instruct}
        \label{fig:step_8b}
    \end{subfigure}
    \hfill
    \begin{subfigure}[b]{0.3\textwidth}
        \centering
        \includegraphics[width=\textwidth]{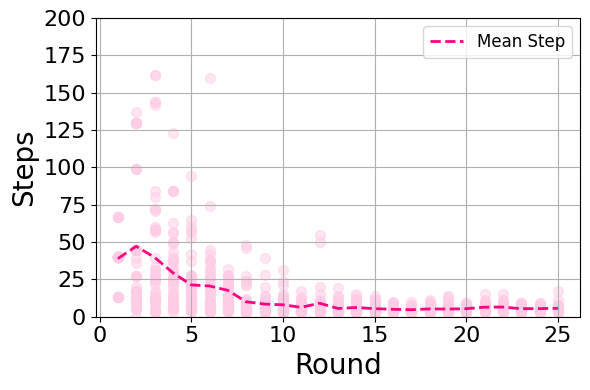}
        \caption{Llama-3.3-70B-Instruct}
        \label{fig:step_70b}
    \end{subfigure}
    \caption{
    Planning step distributions over 25 rounds, aggregated across 50 games played against Llama-3.3-70B-Instruct. }
    \label{fig:step_per_round}
\end{figure*}

Figure~\ref{fig:step_per_round} show the planning depth—measured as the number of steps generated per move—across different LLM architectures. The results indicate that the \text{Llama-3.3-70B-Instruct} model achieves the highest planning depth, with some moves exceeding 250 steps, demonstrating its superior ability to generate long-term sequences. The \text{Llama-3.1-8B-Instruct} follows closely, with a planning capacity reaching around 150 steps. In contrast, the \text{Llama-3.2-3B-Instruct} model exhibits shallower planning, with most of its moves remaining below 100 steps. These findings suggest that larger models possess not only better language understanding but also more advanced planning capabilities in complex, rule-based environments like Ô Ăn Quan.

Interestingly, we also observe a consistent decline in planning depth as the game progresses. Across all three models, early rounds (Rounds 1–5) tend to exhibit the highest number of steps per move, often exceeding 100 steps for the larger models. However, as the game advances into the mid (Rounds 6–15) and late stages (Rounds 16–25), the number of generated steps significantly decreases. This trend suggests that LLMs rely more on long-term planning at the beginning of the game, where the board is in a more stable and predictable state. In contrast, later rounds introduce more variability and constraints, potentially making it harder for models to sustain deep reasoning chains or rendering short-term tactics more effective.

% \subsection{Temporal Dynamics in Decision-Making}
\subsection{Qualitative Analysis of Model Reasoning}

During each game, we collect the step-by-step reasoning produced by the agents. These reasoning traces are then classified using the Gemini-2.0-Flask~\cite{team2023gemini} model in a zero-shot setting into one of three categories: \textit{SHORT\_TERM\_GAIN}, \textit{LONG\_TERM\_STRATEGY}, or \textit{AMBIGUOUS} ( See detail in Appendix~\ref{appendix:prompt}). 

\begin{table}[ht]
\centering
\caption{Distribution of Reasoning Types Across LLM Agents (in \%). 
STG: Short-Term Gain, LTS: Long-Term Strategy, AMB: Ambiguous. The \textbf{best} overall results are in bold, and the \underline{second best} are underlined.}
\renewcommand{\arraystretch}{1.3}
\begin{tabular}{|l|c|c|c|}
\hline
\textbf{Model} & \textbf{STG (\%)} & \textbf{LTS (\%)} & \textbf{AMB (\%)} \\
\hline
Llama-3.2-3B-Instruct & \textbf{70.18} & 23.42 & \textbf{6.40} \\
Llama-3.1-8B-Instruct & \underline{41.87} & \underline{57.67} & 0.46 \\
Llama-3.3-70B-Instruct & 34.50 & \textbf{64.86} & \underline{0.64} \\
\hline
\end{tabular}
\label{tab:reason_type_distribution}
\end{table}

Table~\ref{tab:reason_type_distribution} reveals distinct patterns in reasoning preferences across different LLM models. The smaller \text{Llama-3.2-3B-Instruct} model heavily favors short-term gain reasoning (70.18\%), indicating a tendency toward opportunistic or greedy tactics. In contrast, the \text{Llama-3.3-70B-Instruct} model demonstrates a strong preference for long-term strategic reasoning (64.86\%) with a significantly lower reliance on short-term gain (34.50\%). \text{Llama-3.1-8B-Instruct} shows the most balanced behavior, with over half of its reasoning focused on long-term strategy (57.67\%) and a relatively low ambiguous rate (0.46\%). These results suggest that model size may correlate with increased planning depth and reduced impulsiveness in multi-step decision-making.

% \begin{figure*}[ht]
%     \centering
%     \begin{subfigure}[b]{0.28\textwidth}
%         \centering
%         \includegraphics[width=\textwidth]{figures/reason3b.png}
%         \caption{Llama-3.2-3B-Instruct}
%         \label{fig:step_3b}
%     \end{subfigure}
%     \hfill
%     \begin{subfigure}[b]{0.28\textwidth}
%         \centering
%         \includegraphics[width=\textwidth]{figures/reason8b.png}
%         \caption{Llama-3.1-8B-Instruct}
%         \label{fig:step_8b}
%     \end{subfigure}
%     \hfill
%     \begin{subfigure}[b]{0.28\textwidth}
%         \centering
%         \includegraphics[width=\textwidth]{figures/reason70b.png}
%         \caption{Llama-3.3-70B-Instruct}
%         \label{fig:step_70b}
%     \end{subfigure}

%     \caption{Reasoning type distributions over 25 rounds, aggregated across 50 games played against Llama-3.3-70B-Instruct.}
%     \label{fig:reason_dynamics}
% \end{figure*}

\begin{figure*}[ht]
    \centering
    \includegraphics[width=\textwidth]{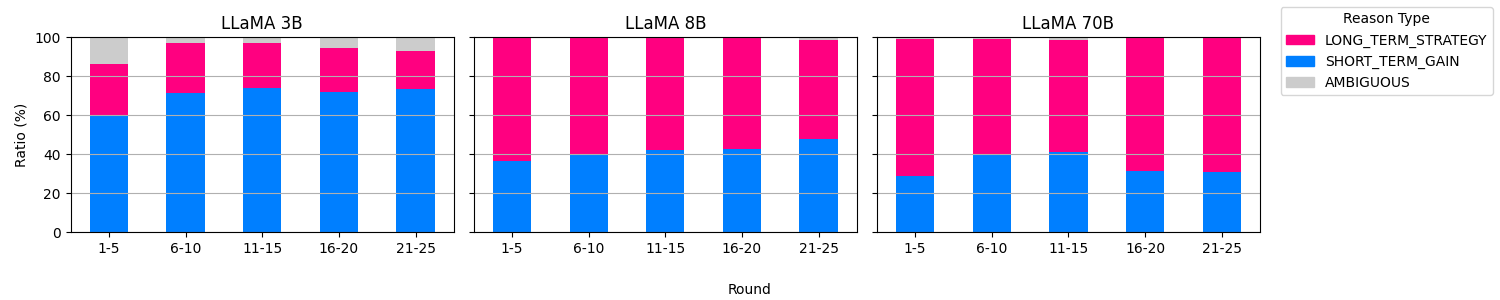}
    \caption{Reasoning type distributions over 25 rounds, aggregated across 50 games played against Llama-3.3-70B-Instruct.}
    \label{fig:reason_dynamics}
\end{figure*}

Figure~\ref{fig:reason_dynamics} illustrates how different reasoning types evolve across game rounds for three LLMs. The \text{Llama-3.3-70B-Instruct} model shows a dominant preference for long-term strategy, consistently maintaining it above 45–60\% throughout most rounds. In contrast, short-term gain reasoning is more prevalent in the 3B and 8B models. Specifically, \text{Llama-3.2-3B-Instruct} exhibits a clear bias towards short-term decisions, peaking at over 75\% from mid to late game, while long-term reasoning remains minimal. The \text{Llama-3.1-8B-Instruct} model presents a more balanced distribution in early rounds but shows a shift toward short-term gain in the final phase. Notably, ambiguous reasoning decreases in higher-capacity models, suggesting that larger models generate more coherent and consistent strategic justifications.

\section{Conclusion}

While originally a traditional Vietnamese folk game, Ô Ăn Quan has been transformed in this work into a structured testbed for evaluating the planning and decision-making capabilities of large language models (LLMs). Its blend of turn-based strategy, dynamic token interactions, and rule constraints provides a rich environment to observe how LLMs reason, adapt, and act.

Our findings highlight that larger models (e.g., Llama-3.3-70B) exhibit superior planning depth, especially in early rounds, but all models show a decline in reasoning steps toward later game phases. Balanced strategies tend to yield higher win rates, while risk-heavy or overly defensive styles underperform.

In future, we hope to see more research that uses traditional or simulated games to study LLM behavior. Extending this idea to other cultural games could help evaluate how well models can plan, adapt, and interact in complex, dynamic environments.

\section*{Limitations}
This study evaluates LLMs in a zero-shot setting without fine-tuning or reinforcement learning. The game environment is simplified, and evaluation is based on reasoning labels instead of full gameplay outcomes. While prompts help guide decisions, the models often struggle with multi-step planning and complex capture chains, limiting their strategic depth.

\section*{Acknowledgment}
We sincerely appreciate the insightful comments and constructive feedback provided by the anonymous reviewers. This research is funded by University of Information
Technology-Vietnam National University Ho Chi Minh City
under grant number D4-2025-05.

\bibliographystyle{IEEEtran}
\bibliography{references}

\begin{thebibliography}{10}
\providecommand{\url}[1]{#1}
\csname url@samestyle\endcsname
\providecommand{\newblock}{\relax}
\providecommand{\bibinfo}[2]{#2}
\providecommand{\BIBentrySTDinterwordspacing}{\spaceskip=0pt\relax}
\providecommand{\BIBentryALTinterwordstretchfactor}{4}
\providecommand{\BIBentryALTinterwordspacing}{\spaceskip=\fontdimen2\font plus
\BIBentryALTinterwordstretchfactor\fontdimen3\font minus \fontdimen4\font\relax}
\providecommand{\BIBforeignlanguage}[2]{{%
\expandafter\ifx\csname l@#1\endcsname\relax
\typeout{** WARNING: IEEEtran.bst: No hyphenation pattern has been}%
\typeout{** loaded for the language `#1'. Using the pattern for}%
\typeout{** the default language instead.}%
\else
\language=\csname l@#1\endcsname
\fi
#2}}
\providecommand{\BIBdecl}{\relax}
\BIBdecl

\bibitem{brown2020language}
T.~Brown, B.~Mann, N.~Ryder, M.~Subbiah, J.~D. Kaplan, P.~Dhariwal, A.~Neelakantan, P.~Shyam, G.~Sastry, A.~Askell \emph{et~al.}, ``Language models are few-shot learners,'' \emph{Advances in neural information processing systems}, vol.~33, pp. 1877--1901, 2020.

\bibitem{touvron2023llama}
H.~Touvron, T.~Lavril, G.~Izacard, X.~Martinet, M.-A. Lachaux, T.~Lacroix, B.~Rozi{\`e}re, N.~Goyal, E.~Hambro, F.~Azhar \emph{et~al.}, ``Llama: Open and efficient foundation language models,'' \emph{arXiv preprint arXiv:2302.13971}, 2023.

\bibitem{touvron2023llama2}
H.~Touvron, L.~Martin, K.~Stone, P.~Albert, A.~Almahairi, Y.~Babaei, N.~Bashlykov, S.~Batra, P.~Bhargava, S.~Bhosale \emph{et~al.}, ``Llama 2: Open foundation and fine-tuned chat models,'' \emph{arXiv preprint arXiv:2307.09288}, 2023.

\bibitem{grattafiori2024llama}
A.~Grattafiori, A.~Dubey, A.~Jauhri, A.~Pandey, A.~Kadian, A.~Al-Dahle, A.~Letman, A.~Mathur, A.~Schelten, A.~Vaughan \emph{et~al.}, ``The llama 3 herd of models,'' \emph{arXiv preprint arXiv:2407.21783}, 2024.

\bibitem{team2023gemini}
G.~Team, R.~Anil, S.~Borgeaud, J.-B. Alayrac, J.~Yu, R.~Soricut, J.~Schalkwyk, A.~M. Dai, A.~Hauth, K.~Millican \emph{et~al.}, ``Gemini: a family of highly capable multimodal models,'' \emph{arXiv preprint arXiv:2312.11805}, 2023.

\bibitem{zhang2023prompting}
B.~Zhang, B.~Haddow, and A.~Birch, ``Prompting large language model for machine translation: A case study,'' in \emph{International Conference on Machine Learning}.\hskip 1em plus 0.5em minus 0.4em\relax PMLR, 2023, pp. 41\,092--41\,110.

\bibitem{kocmi2023large}
T.~Kocmi and C.~Federmann, ``Large language models are state-of-the-art evaluators of translation quality,'' \emph{arXiv preprint arXiv:2302.14520}, 2023.

\bibitem{zhang2024comprehensive}
Y.~Zhang, H.~Jin, D.~Meng, J.~Wang, and J.~Tan, ``A comprehensive survey on process-oriented automatic text summarization with exploration of llm-based methods,'' \emph{arXiv preprint arXiv:2403.02901}, 2024.

\bibitem{lewis2020retrieval}
P.~Lewis, E.~Perez, A.~Piktus, F.~Petroni, V.~Karpukhin, N.~Goyal, H.~K{\"u}ttler, M.~Lewis, W.-t. Yih, T.~Rockt{\"a}schel \emph{et~al.}, ``Retrieval-augmented generation for knowledge-intensive nlp tasks,'' \emph{Advances in neural information processing systems}, vol.~33, pp. 9459--9474, 2020.

\bibitem{huang2024far}
J.-t. Huang, E.~J. Li, M.~H. Lam, T.~Liang, W.~Wang, Y.~Yuan, W.~Jiao, X.~Wang, Z.~Tu, and M.~R. Lyu, ``How far are we on the decision-making of llms? evaluating llms' gaming ability in multi-agent environments,'' \emph{arXiv preprint arXiv:2403.11807}, 2024.

\bibitem{sato2024implementation}
T.~Sato, S.~Ozaki, and D.~Yokoyama, ``An implementation of werewolf agent that does not truly trust llms,'' \emph{arXiv preprint arXiv:2409.01575}, 2024.

\bibitem{lan2023llm}
Y.~Lan, Z.~Hu, L.~Wang, Y.~Wang, D.~Ye, P.~Zhao, E.-P. Lim, H.~Xiong, and H.~Wang, ``Llm-based agent society investigation: Collaboration and confrontation in avalon gameplay,'' \emph{arXiv preprint arXiv:2310.14985}, 2023.

\bibitem{wei2025exploring}
C.~Wei, J.~Chen, and J.~Xu, ``Exploring large language models for word games: Who is the spy?'' \emph{arXiv preprint arXiv:2503.15235}, 2025.

\bibitem{von2007theory}
J.~Von~Neumann and O.~Morgenstern, ``Theory of games and economic behavior: 60th anniversary commemorative edition,'' in \emph{Theory of games and economic behavior}.\hskip 1em plus 0.5em minus 0.4em\relax Princeton university press, 2007.

\bibitem{silver2016mastering}
D.~Silver, A.~Huang, C.~J. Maddison, A.~Guez, L.~Sifre, G.~Van Den~Driessche, J.~Schrittwieser, I.~Antonoglou, V.~Panneershelvam, M.~Lanctot \emph{et~al.}, ``Mastering the game of go with deep neural networks and tree search,'' \emph{nature}, vol. 529, no. 7587, pp. 484--489, 2016.

\bibitem{silver2018general}
D.~Silver, T.~Hubert, J.~Schrittwieser, I.~Antonoglou, M.~Lai, A.~Guez, M.~Lanctot, L.~Sifre, D.~Kumaran, T.~Graepel \emph{et~al.}, ``A general reinforcement learning algorithm that masters chess, shogi, and go through self-play,'' \emph{Science}, vol. 362, no. 6419, pp. 1140--1144, 2018.

\bibitem{silver2017mastering}
D.~Silver, J.~Schrittwieser, K.~Simonyan, I.~Antonoglou, A.~Huang, A.~Guez, T.~Hubert, L.~Baker, M.~Lai, A.~Bolton \emph{et~al.}, ``Mastering the game of go without human knowledge,'' \emph{nature}, vol. 550, no. 7676, pp. 354--359, 2017.

\bibitem{rabinowitz2018machine}
N.~Rabinowitz, F.~Perbet, F.~Song, C.~Zhang, S.~A. Eslami, and M.~Botvinick, ``Machine theory of mind,'' in \emph{International conference on machine learning}.\hskip 1em plus 0.5em minus 0.4em\relax PMLR, 2018, pp. 4218--4227.

\bibitem{wei2022chain}
J.~Wei, X.~Wang, D.~Schuurmans, M.~Bosma, F.~Xia, E.~Chi, Q.~V. Le, D.~Zhou \emph{et~al.}, ``Chain-of-thought prompting elicits reasoning in large language models,'' \emph{Advances in neural information processing systems}, vol.~35, pp. 24\,824--24\,837, 2022.

\bibitem{yao2023react}
S.~Yao, J.~Zhao, D.~Yu, N.~Du, I.~Shafran, K.~Narasimhan, and Y.~Cao, ``React: Synergizing reasoning and acting in language models,'' in \emph{International Conference on Learning Representations (ICLR)}, 2023.

\bibitem{park2023generative}
J.~S. Park, J.~O'Brien, C.~J. Cai, M.~R. Morris, P.~Liang, and M.~S. Bernstein, ``Generative agents: Interactive simulacra of human behavior,'' in \emph{Proceedings of the 36th annual acm symposium on user interface software and technology}, 2023, pp. 1--22.

\end{thebibliography}

\section{Appendix}

\subsection{Source Code}
The source code implementing the game environment, rule system, and LLM interaction described in this paper is available at: \url{https://github.com/ngwgsang/llama-plays-o-an-quan}.

\subsection{Prompt Settings}
\label{appendix:prompt}

We provide two zero-shot prompts used in the experiments. The first prompt helps the agent decide what move to take based on the game state, rules, and strategy. The second prompt is used to classify the type of reasoning based on the explanation given by the model.

\begin{tcolorbox}[
colframe=black!80!white, 
colback=black!10, 
coltitle=white, 
sharp corners=southwest, 
boxrule=0.8mm, 
fontupper=\ttfamily\small,
width=0.48\textwidth, 
title=Zero-shot Decision-Making Prompt,
]

    You are an intelligent agent playing the traditional Vietnamese game "Ô Ăn Quan".

    ---
    
     **Persona**: 
     
     <persona>

    ---
    
    **Game States**
    After the opponent takes action, here is the current board state: 

    <game\_state>

    ---

    **History**

    My thoughts on the previous round: <history>

    ---
    
    **Game Rules**
    
    <rules>

    ---

    % **Output Format**
    
    % <output\_format>

    % ---
    
    **Task**
    
    Based on the above rules and current game state, think about:
    
    - Which position should you pick to scatter from?
    
    - Which direction to scatter?
    
    - Which move fits your strategy?

\end{tcolorbox}

\begin{tcolorbox}[
colframe=black!80!white, 
colback=black!10, 
coltitle=white, 
sharp corners=southwest, 
boxrule=0.8mm, 
fontupper=\ttfamily\small,
width=0.48\textwidth, 
title=Zero-shot Reasoning-Type Classification Prompt,
]

You are analyzing the reasoning behind a move in the traditional Vietnamese board game Ô Ăn Quan. 
In this game, players distribute peasant tokens across positions and try to capture their opponent's tokens, following complex rules including protection of Mandarin positions (Quan), avoiding Immature Mandarins, and maximizing long-term advantage.
Given the player's reasoning below, classify it into one of the following reason type.
In addition to predicting the label, please transcribe the reasoning and highlight the parts that led to the model's labeling decision using **bold text**.

Reasoning: <think>
\end{tcolorbox}

\end{document}